\documentclass[10pt,twocolumn,letterpaper]{article}

\usepackage[pagenumbers]{cvpr} 

\usepackage[dvipsnames]{xcolor}
\definecolor{cvprblue}{rgb}{0.21,0.49,0.74}
\newcommand{\gr}[1]{\textcolor[rgb]{0.30, 0.65, 0.30}{#1}}
\usepackage[pagebackref,breaklinks,colorlinks,allcolors=cvprblue]{hyperref}

\usepackage{algorithm}
\usepackage{algorithmic}
\usepackage{multirow}
\usepackage{graphicx}

\def\paperID{1033} 
\def\confName{CVPR}
\def\confYear{2025}

\title{Temporal Action Detection Model Compression by Progressive Block Drop}

\author{
Xiaoyong Chen$^{1,2\text{*}}$, Yong Guo$^{3}$\thanks{Equal contribution}, Jiaming Liang$^{2}$, Sitong Zhuang$^{2}$, Runhao Zeng$^{1,2}$\textsuperscript{\textdagger}, Xiping Hu$^{1,4}\thanks{Corresponding author}$ \\
$^1$Guangdong-Hong Kong-Macao Joint Laboratory for Emotional Intelligence and Pervasive Computing,\\Shenzhen MSU-BIT University, $^2$Shenzhen University,\\$^3$South China University of Technology, $^4$Beijing Institute of Technology\\
{\tt\small xiaoyongchencs@gmail.com, zengrh@smbu.edu.cn, huxp@smbu.edu.cn}}

\begin{document}
\maketitle
\begin{abstract}

Temporal action detection (TAD) aims to identify and localize action instances in untrimmed videos, which is essential for various video understanding tasks. However, recent improvements in model performance, driven by larger feature extractors and datasets, have led to increased computational demands. This presents a challenge for applications like autonomous driving and robotics, which rely on limited computational resources. 
While existing channel pruning methods can compress these models, reducing the number of channels often hinders the parallelization efficiency of GPU, due to the inefficient multiplication between small matrices. Instead of pruning channels, we propose a \textbf{Progressive Block Drop} method that reduces model depth while retaining layer width. In this way, we still use large matrices for computation but reduce the number of multiplications. Our approach iteratively removes redundant blocks in two steps: first, we drop blocks with minimal impact on model performance; and second, we employ a parameter-efficient cross-depth alignment technique, fine-tuning the pruned model to restore model accuracy. Our method achieves a 25\% reduction in computational overhead on two TAD benchmarks (THUMOS14 and ActivityNet-1.3) to achieve lossless compression. More critically, we empirically show that our method is orthogonal to channel pruning methods and can be combined with it to yield further efficiency gains.
\end{abstract}

\section{Introduction}
Temporal action detection aims to identify action instances and localize their start and end times in untrimmed videos. This task underpins many video understanding applications, including video question answering~\cite{lei2018tvqa} and video captioning~\cite{gao2017video}. Recent advancements in larger feature extractors and datasets have improved model performance but also significantly increased computational demands~\cite{liu2024end}. However, many TAD applications, such as autonomous driving~\cite{teichmann2018multinet} and robotics~\cite{lynch2023interactive}, operate under limited computational resources. Thus, reducing computational complexity is crucial for deploying TAD in real-world scenarios.

\begin{figure*}[t]
    \centering
    \includegraphics[width=2.\columnwidth]{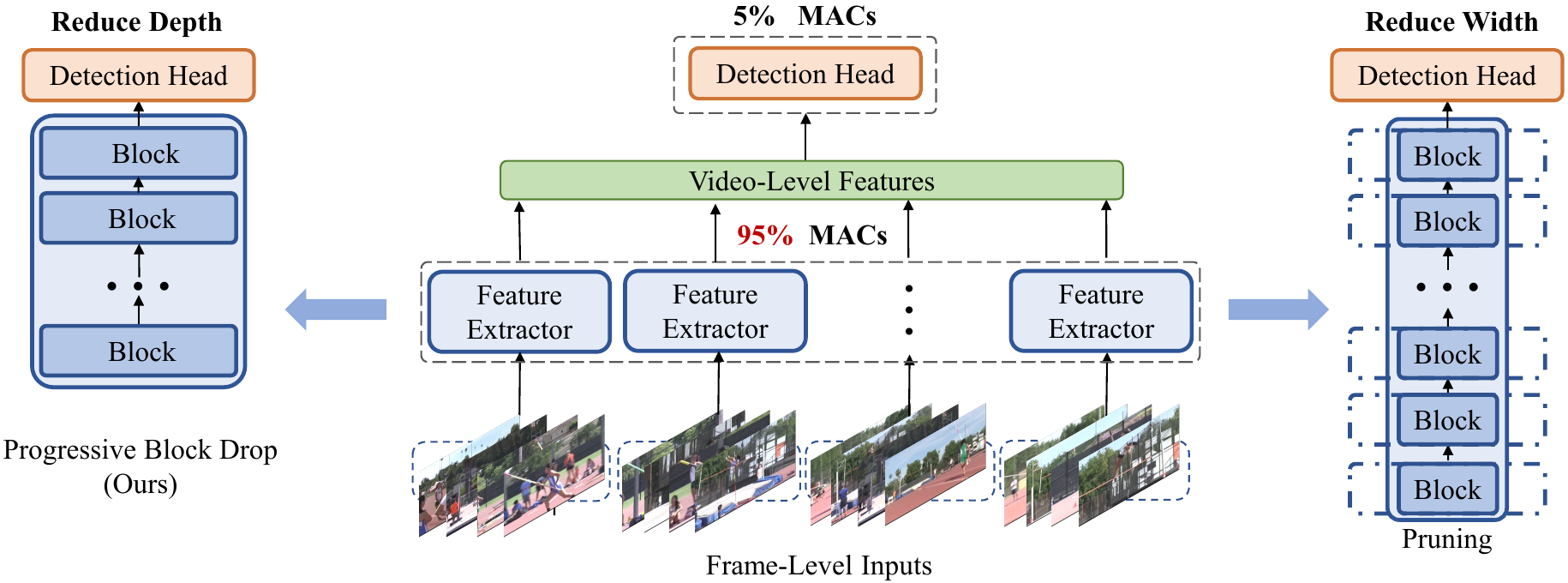}
    \caption{
    Comparisons between reducing depth and reducing width. For a 768-frame video sequence with a resolution of $160^2$, using VideoMAE-S as the feature extractor and Actionformer as the detection head, 95\% of the computational overhead comes from the feature extractor. Most pruning methods reduce the size of layer weight matrices, leading to a slim-and-tall network structure. In contrast, our progressive block drop method reduces network depth, achieving 1.19× faster inference at the same computational cost. These findings, detailed in Section~\ref{sec:Inference Time}, suggest that our approach results in a more hardware-efficient model.
    }
    \label{Fig.pruning_layer_drop} 
    \vspace{-0.4cm}
\end{figure*}

A typical TAD model~\cite{zhang2022actionformer,yang2023basictad} consists of a feature extractor and an action detection head. Due to the long duration of videos, the feature extractor processes video segments individually, and these segment-level features are then combined before being passed to the detection head. As shown in Figure~\ref{Fig.pruning_layer_drop}, our experiments show that the feature extraction phase accounts for 95\% of the total model computation, since it processes many segments, while the detection head processes only a single video-level feature. Thus, reducing the complexity of the feature extractor is key to improving TAD model efficiency.

\label{sec:intro}

Pruning~\cite{molchanov2019pruning, franklelottery, han2015learning, gaodynamic} is a common model compression technique, which removes redundant weights based on their importance. This results in a slim-and-tall network structure, where the model’s hidden layer dimensions are reduced but depth remains unchanged, as illustrated in Figure~\ref{Fig.pruning_layer_drop}. However, studies have shown~\cite{hoefler2021sparsity, ma2021non} that for the same number of parameters, a deep-and-thin network may have slower inference speeds compared to a shallow-and-wide network. The reason for this is that, during GPU inference, operations on larger matrices tend to have higher parallelism and are thus more efficient than multiple operations on smaller matrices.

To explore the feasibility of reducing the architectural depth of TAD models, we conduct preliminary experiments on the THUMOS14 dataset~\cite{jiang2014thumos}, which is widely used for this task. We raise the following key observations from Figure~\ref{Fig.block_drop_effect}: 1) There is a slight discrepancy between the input and output features of certain blocks. 2) Dropping specific blocks results in only a minimal decrease in model performance. 3) Simultaneously dropping multiple blocks leads to a more significant performance drop (Table~\ref{exp:sim_drop}). These results motivate us to investigate TAD model compression from the depth perspective.

In this paper, we propose a multi-step progressive block drop method that effectively reduces network depth, transforming the TAD model into a shallow-and-wide architecture more suitable for parallel GPU matrix computations. Specifically, we compress the model progressively through multiple iterations. Each iteration consists of two stages: 1) block selection and removal, where we evaluate the importance of each block based on a specific metric and remove the least important block; and 2) performance restoration, where we design a parameter-efficient cross-depth alignment strategy. This strategy first integrates LoRA~\cite{hulora} into the attention modules of the remaining blocks and then fine-tunes the pruned model to align with the intermediate features output by the uncompressed model's blocks, effectively compressing the information from multiple blocks into a single block. Additionally, we employ prediction-level alignment to further improve performance. On the THUMOS14 and ActivityNet datasets~\cite{caba2015activitynet}, our method achieves superior performance compared to the uncompressed model, while reducing the model's computational cost (MACs) by 25\%. Furthermore, we demonstrate through experiments that our approach is compatible with common pruning techniques, enabling even further compression of the model. Our main contributions are as follows:
\begin{itemize}
    \item We focus on compressing the TAD model from the perspective of depth. Unlike channel pruning methods that compress the model from the width or the internal layers of blocks, we analyze the redundancy within the TAD model and explore block-level compression as a novel approach. This provides a new angle for compressing and accelerating the TAD model.
    \item To achieve depth compression, we propose a progressive multi-step block drop strategy. In each iteration, a block is automatically removed, and a parameter-efficient training technique is employed to recover the model's performance. On THUMOS14 and ActivityNet, our method reduces the model's MACs by approximately 25\%, while simultaneously outperforming the uncompressed model. Furthermore, we extend our method to other TAD architectures, fine-grained datasets, and additional tasks to validate its generalizability.
    \item We empirically demonstrate that our method is orthogonal to channel pruning techniques and investigate its compatibility with other model compression approaches, resulting in additional efficiency improvements.
\end{itemize}

\section{Related Work}
\label{sec:related_work}
\subsection{Temporal Action Detection} 
Temporal action detection (TAD) can generally be divided into two primary categories~\cite{wang2023temporal}. The first category, two-stage methods, involves generating action proposals followed by classification and boundary refinement~\cite{chao2018rethinking, shou2016temporal, xu2017r, zeng2021graph, zhao2017temporal}. Proposal generation techniques may employ frame- or segment-level classification with subsequent merging of frames or segments belonging to the same action class~\cite{montes2016temporal, nag2022proposal, piergiovanni2019temporal}, while others use specialized proposal generation strategies~\cite{buch2017sst, lin2019bmn}. However, the success of these approaches heavily relies on the quality of the proposals, which has led to the emergence of integrated approaches, also known as one-stage methods. These methods simultaneously handle proposal generation, classification, and/or boundary regression~\cite{huang2019decoupling, lin2017single, tang2023temporalmaxer, yeung2016end}. Key developments in this area include the adoption of anchor-based mechanisms for TAD~\cite{lin2017single} and the shift towards anchor-free frameworks~\cite{lin2021learning, shi2023tridet}, with recent efforts combining the benefits of both anchor-based and anchor-free approaches~\cite{yang2020revisiting}.

In recent years, Transformer-based architectures, which have demonstrated remarkable results in various computer vision tasks, have been applied to TAD~\cite{zhang2022actionformer,tan2021relaxed,wang2021temporal}. Additionally, the development of end-to-end architectures has been explored in various studies~\cite{liu2022empirical, liu2022end, yang2023basictad, liu2024end, zhao2023re2tal}. With the significant achievements of scaling up in large language models~\cite{achiam2023gpt}, recent research~\cite{liu2024end} has utilized scaling up in conjunction with a frame input data processing approach. This combination has not only significantly improved model performance but also reduced computational costs. However, the issue of increased inference time due to computational scaling has become increasingly critical. Therefore, we focus on reducing inference latency through model compression.

\subsection{Model Compression}

Model compression techniques are typically categorized into pruning~\cite{han2015learning}, knowledge distillation~\cite{hinton2015distilling}, model decomposition~\cite{denton2014exploiting}, and quantization~\cite{krizhevsky2012imagenet}. This paper focuses on model pruning, which can be either unstructured~\cite{han2015learning} or structured~\cite{li2016pruning}. Unstructured pruning removes network connections to create sparsity, but it requires specialized hardware for sparse matrix operations~\cite{han2015learning, han2015deep, srinivas2017training}, limiting its practical deployment. Structured pruning, which includes width-reduction methods like channel pruning~\cite{he2017channel, li2016pruning} and filter pruning~\cite{he2019filter}, results in a dense weight matrix, enabling GPU acceleration. However, it may not fully exploit GPU parallelism~\cite{liu2020autocompress}.

Depth-wise pruning techniques~\cite{howard2019searching, tan2019mnasnet} have been used for model acceleration, although they may suffer from insufficient memory occupancy. Various methods, including layer importance evaluation~\cite{chen2018shallowing, elkerdawy2020filter}, parameter optimization~\cite{zhou2021evolutionary, xu2020layer}, and reparameterization~\cite{dror2021layer, fu2022depthshrinker}, can also guide pruning to compress the model from the perspective of depth. 
Previous work~\cite{liu2024updp, zhang2020accelerating} has explored the concept of progressive block dropping; however, there are significant differences between their approaches and ours. Specifically,~\cite{liu2024updp} focuses on compressing and integrating the layers within the block, which is a layer-level pruning technique. The dynamically skipping blocks method proposed in~\cite{zhang2020accelerating} accelerates training by dropping blocks, but during inference, the full model is still employed. Additionally, skipping blocks to reduce the depth of Transformer models~\cite{dong2021attention, michel2019sixteen} often leads to a loss in accuracy. In contrast, our progressive drop block method reduces depth at the block level, improving both inference speed and detection accuracy. Our method can be combined with structured pruning~\cite{fang2023depgraph} and unstructured activation sparsity~\cite{mirzadehrelu, kurtz2020inducing} to further compress the model and accelerate inference while improving performance.

\begin{figure}[t]
    \centering 
    \includegraphics[width=1.\linewidth]{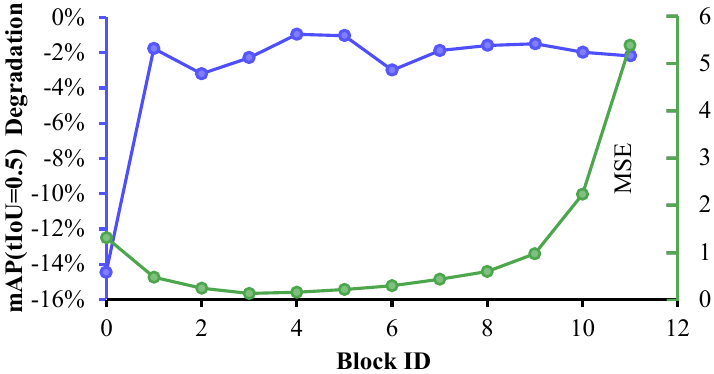}
    \caption{
    Analysis of TAD models at block level on THUMOS14 datasets, using VideoMAE-S.   \textbf{\textcolor{blue}{Blue curve}}:  impact of dropping a single block on the model's detection accuracy, where most cases result in minimal accuracy degradation. \textbf{\gr{Green line}}: MSE of the input and output features for each block. Some blocks exhibit MSE values close to 0, indicating that these blocks are redundant.}
    \label{Fig.block_drop_effect} 
    \vspace{-0.4cm}
\end{figure}

\section{Preliminaries and Motivation}
\label{sec:proposed_method}
In this section, we first introduce the notations and formulations pertinent to temporal action detection (TAD) and then explore the feasibility of reducing the depth of TAD models, as well as some key findings that inspire the design of subsequent methods.

\subsection{TAD Formulation and Notation} 

TAD aims to identify action instances within a video and determine their temporal boundaries. Given an untrimmed video represented as \( V = \{I_t\}_{t=1}^T \), where \( I_t \) denotes the frame at time \( t \), TAD predicts a set of action instances \( \Phi_V = \{\phi_i = (t_{s_i}, t_{e_i}, c_i)\}_{i=1}^N \). Here, \( N \) is the number of action instances; for each instance \( \phi_i \), \( t_{s_i} \) and \( t_{e_i} \) are its starting and ending times, respectively, and \( c_i \) denotes its category.

\begin{figure*}[!t]
    \centering 
    \includegraphics[width=\linewidth]{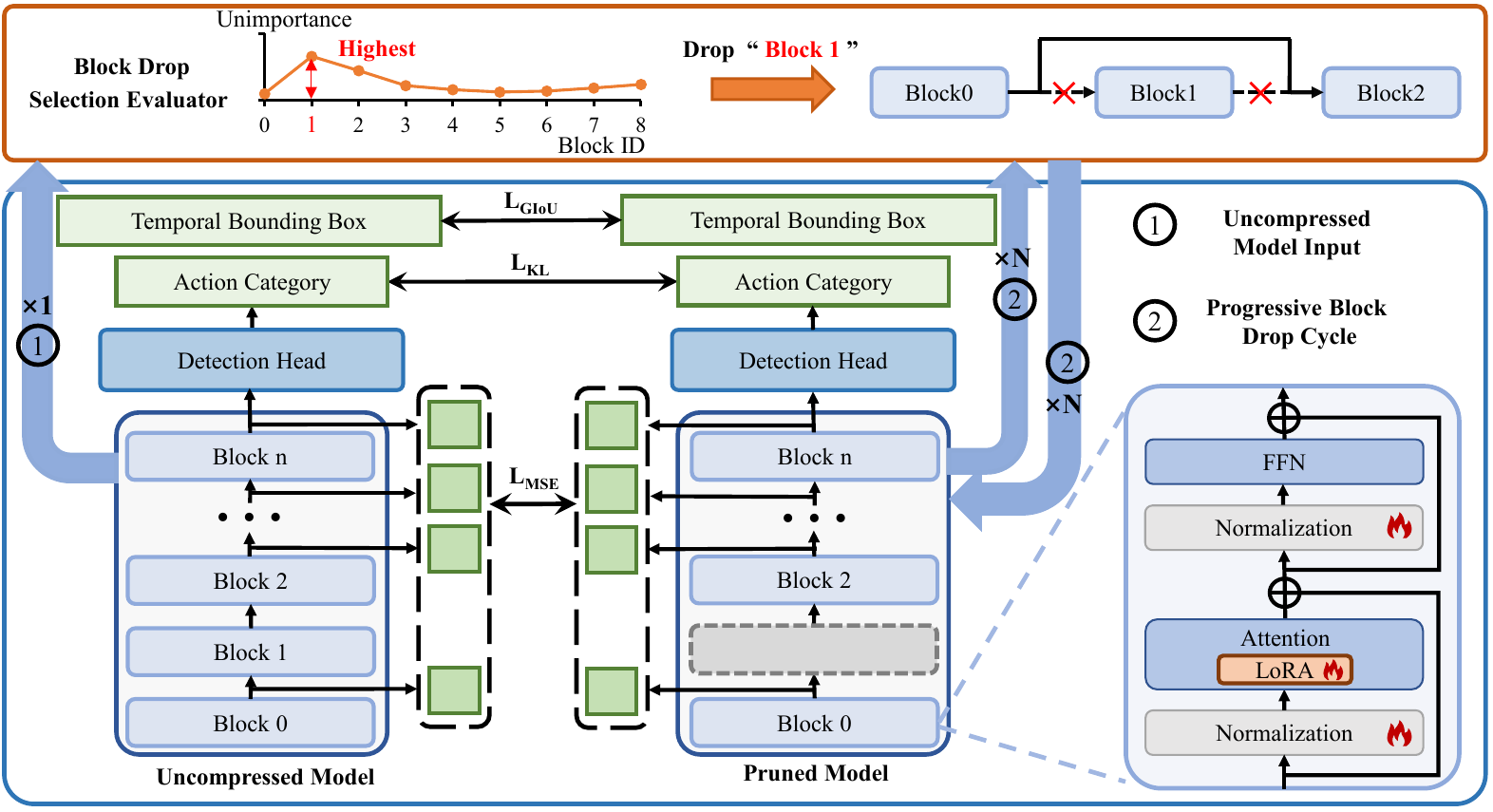}
    \caption{
    The diagram of our progressive block drop method. Our approach adopts a multi-step progressive compression strategy. At each iteration 1) we evaluate the importance of each block and drop the least important block, and 2) we use parameter-efficient tuning techniques, and recover performance by learning from the uncompressed model through feature-level and prediction-level alignment.}
    \label{Fig.progressive_layer_drop} 
    \vspace{-0.4cm}
\end{figure*}

\subsection{Feasibility of Reducing TAD model's Depth}\label{sec:Feasibility Analysis}
We find that there is a severe redundancy in the depth of TAD models. To verify this, we conduct a series of experiments and summarize the key observations as follows.

 \textbf{Observation 1: Most computational overhead lies in the clip-level feature extraction backbone.}
TAD methods typically consist of two main parts: the feature extractor and the detection head. In the feature extraction stage, a sliding window of length \( L \) is processed by a video feature extractor. For instance, features are extracted from frames \(\{ I_{t}, \dots, I_{t+L} \}\), and then the window shifts by \( s \) frames to extract features from frames \( \{I_{t+s}, \dots, I_{t+s+L} \}\). As the feature extraction process is repeated across the entire video sequence, it incurs substantial computational cost, with each window requiring complex operations to extract features. Once the features are extracted, they are concatenated along the temporal dimension, and information is fused across the spatial dimension by averaging the features. These fused features are then passed to the detection head. Ultimately, the volume of data fed into the detection head is much smaller compared to the data processed by the feature extractor. As shown in Figure~\ref{Fig.pruning_layer_drop}, the computational cost of the feature extractor accounts for 95\% of the total forward inference cost of the model. Given the high computational cost of feature extraction, optimizing or reducing the complexity of this stage presents a significant opportunity for improving model efficiency. Therefore, to reduce the overall model computation, the primary focus should be on compressing the feature extractor.

\textbf{Observation 2: There exist blocks that have a minor discrepancy between their input and output.}
Additionally, by analyzing the feature differences between consecutive blocks, as depicted in Figure~\ref{Fig.block_drop_effect}, we find that some blocks exhibit very similar representation capabilities, suggesting redundancy in the model. These findings highlight the feasibility of applying the block drop method to reduce computational complexity without sacrificing performance.

\textbf{Observation 3: The removal of certain blocks has a negligible impact on TAD performance.}
To investigate the potential for model compression, we first assess the feasibility of dropping blocks by examining the impact of removing individual blocks on model performance. As shown in Figure~\ref{Fig.block_drop_effect}, our experiments reveal that removing certain blocks led to only minimal performance degradation, indicating that block drop could be a viable approach.

\textbf{Observation 4: Simultaneously removing multiple blocks at a time results in a significant performance drop.}\label{sec:simultaneously removing multiple blocks}
In our preliminary exploration, we attempt to directly drop multiple blocks and train the model to observe whether it could recover its performance. We randomly select three blocks to drop all at once and then attempt fine-tuning. As shown in Table~\ref{exp:sim_drop}, the best performance achieved is 0.67\% lower than the uncompressed model, exhibiting a noticeable performance drop, which is an undesirable outcome in practical applications. Therefore, we adopt a multi-step progressive block dropping strategy rather than removing multiple blocks at once. This approach allows the information from the removed blocks to be more effectively compressed into the pruned model, resulting in better performance.

\section{Progressive Block Drop}
In this paper, we design a compression method specifically tailored for TAD models. Drawing from our analysis in Section~\ref{sec:Feasibility Analysis}, our focus is on reducing the computational cost of the feature extraction backbone while maintaining or even enhancing performance. To achieve this, we propose a multi-step progressive block-level dropping method, where each step comprises two stages. First, the \textbf{Block Selection Evaluator} automatically selects the blocks to be dropped, resulting in a subnet. Then, \textbf{Cross-Depth Alignment} enables parameter-efficient learning from the uncompressed model to maintain TAD performance. By iteratively applying these steps, we progressively obtain a pruned TAD backbone that can be directly deployed for inference.

\subsection{Block Selection Evaluator}
Experimental results presented in Section~\ref{sec:simultaneously removing multiple blocks} demonstrate that simultaneously removing multiple blocks significantly degrades the performance of TAD. Inspired by prior work on curriculum learning~\cite{bengio2009curriculum, wang2021survey}, we propose a progressive block-drop strategy. However, selecting the appropriate blocks to drop is a non-trivial task. To address this challenge, we introduce a block selection evaluator $f_E$ to select the blocks to be dropped automatically. 

Without loss of generality, let \( M_0 \) denote the initial uncompressed model trained on the training set, and let \( M_{t-1} \) represent the model input to $f_E$ at iteration \( t \). We denote \( B_{t-1} = \{b_1, b_2, \dots, b_K\} \) as the set of all blocks within \( M_{t-1} \), where \( K \) is the total number of blocks. At iteration \( t \), the collection of sub-networks resulting from dropping each single block is denoted as:
\begin{equation}
 \mathcal{S}_t = \{M^{sub}_{t,k} = M_{t-1} \setminus b_k |\ b_k \in B_{t-1} \}.
\end{equation}
To automatically select blocks for pruning, we employ the evaluation function \( f_E \) to assess each subnet. By measuring the performance difference between each subnet and the uncompressed TAD model \( M_0 \) on the training set, \( f_E \) outputs an importance score for each subnet; the smaller the performance difference, the higher the importance score assigned. This process can be formulated as:
\begin{equation}
\mathcal{P}_t = f_E(\mathcal{S}_t, D_{\text{train}}).
\end{equation}
From the set of importance scores \( \mathcal{P}_t \), we select the subnet \( M^{sub}_{t,*} \) with the highest importance score \( p^* \), resulting in the pruned model \( M_t \) at step \( t \). Note that \( f_E \) can also compute the difference between the input and output features of each removed block or the training loss of each subnet on the training set; in both cases, a smaller difference corresponds to a higher importance score. In our experiments (Section~\ref{sec:different metrics}), we find that selecting subnets based on the mAP metric from the training set yields the best results.

\subsection{Parameter-Efficient Learning from Uncompressed Model for Performance Recovery}\label{sec:Performance Recovery}

After removing one block from the TAD model, a central challenge arises in ensuring that the performance of the reduced-depth model remains comparable to that of the original, uncompressed model (see Section~\ref{sec:effect of FT} of appendix). However, training the TAD model on untrimmed videos using full-parameter fine-tuning still demands a substantial amount of GPU memory, as discussed in \cite{liu2024end}. For example, in our experiments with the VideoMAE-L model we used, setting the batch size to 1 already consumes 13.1 GB of memory. Therefore, we introduce LoRA to reduce memory consumption during training, with detailed experimental results provided in Section~\ref{sec:impact of LoRA} of the appendix.

Before training, given the pruned model \( M_t(\theta_t)\) at step \( t \), where \( \theta_t \) represents its parameters, we first follow the LoRA method~\cite{hulora, chenlonglora} to add trainable parameters \( \theta_{\text{LoRA}} \) to the attention layer each block, resulting in the model \( \hat{M}_t(\theta_t, \theta_{\text{LoRA}}) \). This approach offers two main benefits: first, during subsequent training, only \( \theta_{\text{LoRA}} \) is trained, significantly reducing memory consumption; second, after training, \( \theta_{\text{LoRA}} \) can be merged with \( \theta_t \), thereby keeping the TAD model's structure unchanged and incurring no additional inference computational overhead.

\begin{algorithm}[!t]
\caption{Proposed Progressive Block Drop Method}
\begin{algorithmic}[]

\REQUIRE Uncompressed model $M_0$, model $M_{t-1}$ at step $t$, blocks $B_{t-1} = \{b_1, b_2, \dots, b_K\}$ within $M_{t-1}$, training set $D_{\text{train}}$.
\STATE // \textbf{Progressive block drop cycle}
\WHILE{Performance of $M_{t-1} \geq M_0$}
    \STATE // \textbf{Stage 1: select one block to drops}
    \STATE Obtain sub-nets by dropping each block:
    
    $\mathcal{S}_t = \left\{ M^{\text{sub}}_{t,k} = M_{t-1} \setminus b_k \ \big| \ b_k \in B_{t-1} \right\}$
    
    \STATE Evaluate each sub-net using the evaluator $f_E$:
    
    $\mathcal{P}_t = f_E(\mathcal{S}_t, D_{\text{train}})$
    
    \STATE Select the subnet with the highest performance:
    
    $M_{t} = \arg\max(\mathcal{P}_t)$
    \STATE // \textbf{Stage 2: performance recovery}
    \STATE  Insert LoRA modules into $M_t$ for training:
    
    $M_t(\theta) \rightarrow \hat{M_t}(\theta, \theta_{\text{LoRA}})$
    
    \STATE Train the model with Eqn. (\ref{eq.total loss})
    
    \STATE  Merge LoRA parameters into the model:
    
    $\theta_t \leftarrow \theta_t + \theta_{\text{LoRA}}'$
\ENDWHILE
\end{algorithmic}
\label{Alg:dld}
\end{algorithm}

\noindent \textbf{Cross-Depth Alignment.}
The structural differences between the pruned and uncompressed models—resulting from block dropping—pose a significant challenge for recovering performance. More critically, we empirically observed that directly supervising the training of \( \hat{M}_t \) with ground truth fails to restore performance (see Section E.5). In other words, relying solely on prediction-level supervision is inadequate for restoring performance. To address this issue, we shift our attention to feature-level supervision and propose a cross-depth alignment strategy.

Specifically, if we consider three adjacent blocks, \( \{b_{i-1}, b_i, b_{i+1}\} \), as a group, when \( b_i \) is dropped, the input to \( b_{i+1} \) changes (i.e., it becomes the output of \( b_{i-1} \)). To maintain performance, it is necessary to: 1) ensure that the input feature distribution of \( b_{i+1}^p \) in the pruned model \( \hat{M}_t \) aligns with that of \( b_i^u \) in the uncompressed model \( M_0 \); and 2) ensure that the output distribution of the \( (i+1) \)-th block in \( \hat{M}_t \) aligns with that of \( M_0 \).
In our implementation, after the \( i \)-th block is dropped, we align the outputs of the \( (i-1) \)-th and \( (i+1) \)-th blocks ($f_{b_{i-1}}$ and $f_{b_{i+1}}$) between the pruned and uncompressed models, thereby compressing the information from multiple blocks into a single block. For the features of other blocks, we similarly apply block-wise alignment with the \( M_0 \). The cross-depth alignment loss can be expressed as follows:
\begin{equation}
\mathcal{L}_{f} = \frac{1}{I-1}\sum_{m=1,m\neq i}^{I} (f^{M_0}_{b_m} - f^{\hat{M}_t}_{b_m})^{2}.
\end{equation}

\noindent \textbf{Action classification and Localization Alignment.} The TAD model generates two distinct types of predictions: action categories and temporal bounding boxes. We also align the predictions of the pruned model \( \hat{M}_t \) with those of the uncompressed model \( M_0 \). 

\textbf{1) Action classification alignment:} To align the action class predictions, we employ the Kullback-Leibler (KL) divergence loss \( \mathcal{L_{\text{KL}}} \)~\cite{kullback1951information}, which measures the divergence between the action class scores of the uncompressed model \( M_0 \) and the pruned model \( \hat{M}_t \). The alignment loss is defined as:
\begin{equation}
\mathcal{L}_{pc} = \mathcal{L}_\text{KL}(p(z_{M_0}), p(z_{\hat{M}_t})),
\end{equation}
where \( z_{M_0} \) and \( z_{\hat{M}_t} \) denote the logits of the uncompressed and pruned models, respectively, and \( p(z_{M_0}) \) and \( p(z_{\hat{M}_t}) \) represent the class probabilities derived from the logits via the softmax function. 

\textbf{2) Boundary regression alignment:} For aligning the boundary predictions, we compute the Generalized Intersection over Union (GIoU) loss \( \mathcal{L}_{\text{GIoU}} \)~\cite{rezatofighi2019generalized} between the predicted temporal bounding boxes of the uncompressed model \( M_0 \) and the pruned model \( \hat{M}_t \). The alignment loss for boundary regression is:
\begin{equation}
\mathcal{L}_{pr} = \frac{1}{N_p}\sum_{i=1}^{N_p}{\mathcal{L}_{\text{GIoU}}(D^{M_0}_{i}, D^{\hat{M}_t}_{i})},
\end{equation}
where \( N_p \) denotes the number of predicted positive temporal bounding boxes, and \( D^{M_0}_{i} \) and \( D^{\hat{M}_t}_{i} \) represent the \( i \)-th predicted temporal bounding boxes for models \( M_0 \) and \( \hat{M}_t \), respectively. This dual alignment ensures that the pruned model learns both the correct action classification and the accurate temporal boundaries, which are crucial for TAD.

\noindent \textbf{Total Loss Function.} We incorporate the alignment losses, along with the commonly used classification loss \( \mathcal{L}_{\text{cls}} \) and regression loss \( \mathcal{L}_{\text{reg}} \) in TAD~\cite{liu2024end}, as follows:
\begin{equation}\label{eq.total loss}
\mathcal{L}_{total} = \mathcal{L}_{pc} + \mathcal{L}_{pr} + \mathcal{L}_{f} + \mathcal{L}_{cls} + \mathcal{L}_{reg}.
\end{equation}
After training, the LoRA parameters \( \theta_{\text{LoRA}}' \) are integrated with the pruned model as:
\begin{equation}
\theta_t \leftarrow \theta_t + \theta_{\text{LoRA}}'.
\end{equation}
The updated model \( M_t(\theta_t) \) is then input into the block drop selection evaluator at the \( t+1 \) step for the next iteration, continuing until the performance of the pruned model is no longer recoverable to that of the uncompressed model. Ultimately, the compressed TAD model is deployed, offering a reduced number of parameters and faster inference speed.

\section{Experiments}
\label{sec:experiments}
In the experimental section, we conduct the following experiments: 1) Sections~\ref{exp:apply block drop method} and~\ref{exp:extended eval} evaluate the effectiveness of the proposed method; 2) Section~\ref{sec:compatibility} validates compatibility with other compression models; 3) Section~\ref{sec:Inference Time} compares inference speed with pruning methods.

\subsection{Datasets and Evaluation Metric}\label{sec:dataset and metrics}

\textbf{THUMOS14~\cite{jiang2014thumos}} is a prominent benchmark dataset in TAD research. It consists of a training set with 200 videos and a test set with 213 videos, covering 20 action categories. This task is notably challenging due to the presence of more than 15 separate action instances per video, with background scenes occupying 71\% of the frames.

\noindent\textbf{ActivityNet-1.3~\cite{caba2015activitynet}} is a large-scale benchmark for TAD in untrimmed videos, containing approximately 20,000 videos across 200 activity classes, totaling over 600 hours of content. The dataset is divided into three subsets: 10,024 videos for training, 4,926 for validation, and 5,044 for testing, with an average of 1.65 action instances per video.

\noindent\textbf{FineAction}~\cite{liu2022fineaction} is a benchmark dataset for temporal action localization, containing 103K instances across 106 action categories in 17K untrimmed videos. It presents both opportunities and challenges for TAD due to its fine-grained action classes, dense annotations, and frequent co-occurrence of multiple action classes.

\noindent\textbf{Mean Average Precision (mAP)}. A predicted temporal bounding box is deemed accurate if its temporal Intersection over Union (tIoU) with the corresponding ground truth instance exceeds a specific threshold and matches the instance’s category. For THUMOS14, the tIoU thresholds range from $\{0.3, 0.4, 0.5, 0.6, 0.7\}$, with mAP@tIoU=0.5 reported for comparison. In contrast, on ActivityNet-1.3 and FineAction, tIoU thresholds are sampled from $\{0.5 : 0.05 : 0.95\}$, and we report the mean mAP over tIoU thresholds from 0.5 to 0.95, in increments of 0.05.

\begin{table*}[t]
\caption{
Evaluation of our method in terms of computational complexity, inference time and mAP with VideoMAE-S as the backbone. Our approach reduces the computational cost by approximately 25\%, achieves around 1.3x inference speedup, and improves the performance.
 }
 \vspace{-0.2cm}
\tabcolsep=3pt
\centering
\footnotesize
\begin{tabular}{ccccccc}
\hline
\multirow{2}{*}{\begin{tabular}[c]{@{}c@{}}Model\\ (Drop \#Block)\end{tabular}} & \multicolumn{3}{|c}{THUMOS14}                         & \multicolumn{3}{|c}{ActivityNet-1.3}                  \\ \cline{2-7} 
 & \multicolumn{1}{|c}{MACs (G)} & \multicolumn{1}{c}{Inference Time (ms)} & mAP (tIoU=0.5) & \multicolumn{1}{|c}{MACs (G)} & \multicolumn{1}{c}{Inference Time (ms)} & mAP (Avg) \\ \hline
Baseline           & \multicolumn{1}{|c}{286.3} & \multicolumn{1}{c}{104.9} & \multicolumn{1}{c}{70.43} & \multicolumn{1}{|c}{274.1} & \multicolumn{1}{c}{103.8} & \multicolumn{1}{c}{37.75} \\ 
 1            & \multicolumn{1}{|c}{263.5\textbf{{{ (92.0\%)}}}} & \multicolumn{1}{c}{98.4 \textbf{{{(93.8\%)}}}} & 71.06 \textbf{{{($\uparrow$ 0.63)}}} & \multicolumn{1}{|c}{251.3 \textbf{{{(91.7\%)}}}} & \multicolumn{1}{c}{95.1 \textbf{{{(91.6\%)}}}} & 37.94 \textbf{{{($\uparrow$ 0.19)}}}\\ 
 2            & \multicolumn{1}{|c}{240.8 \textbf{{{(84.1\%)}}}} & \multicolumn{1}{c}{89.8\textbf{{{ (85.6\%)}}}} & 71.37 \textbf{{{($\uparrow$ 0.94)}}}& \multicolumn{1}{|c}{228.5 \textbf{{{(83.4\%)}}}} & \multicolumn{1}{c}{86.8 \textbf{{{(83.6\%)}}}} & 37.81 \textbf{{{($\uparrow$ 0.06)}}}\\ 
 3            & \multicolumn{1}{|c}{218.0 \textbf{{{(76.1\%)}}}} & \multicolumn{1}{c}{81.2 \textbf{{{(77.4\%)}}}} & 70.47 \textbf{{{($\uparrow$ 0.04)}}}& \multicolumn{1}{|c}{205.8 \textbf{{{(75.1\%)}}}} & \multicolumn{1}{c}{79.6 \textbf{{{(76.7\%)}}}} & 37.77 \textbf{{{($\uparrow$ 0.02)}}}\\ 
  4            & \multicolumn{1}{|c}{195.2 \textbf{{{(68.2\%)}}}} & \multicolumn{1}{c}{73.6 \textbf{{{(70.2\%)}}}} & 69.65 \textbf{($\downarrow$ 0.78)}& \multicolumn{1}{|c}{183.0 \textbf{{{(66.8\%)}}}} & \multicolumn{1}{c}{72.1 \textbf{{{(69.5\%)}}}} & 37.72 \textbf{{{($\downarrow$ 0.03)}}}\\ \hline
\end{tabular}\label{exp:block drop small}
\vspace{-0.3cm}
\end{table*}

\begin{table}[t]
\caption{
Evaluation of our method on deeper networks in terms of computational complexity, inference time and mAP with VideoMAE-L (24 blocks) as the backbone. Our method is generally effective and larger models may benefit more from it.
}\vspace{-0.3cm}
\tabcolsep=3pt
\resizebox{\columnwidth}{!}{
\normalsize
\begin{tabular}{ccccc}
\hline
Drop \#Block & MACs (G) & \multicolumn{1}{c}{Inference Time (ms)} & \multicolumn{1}{c}{mAP (tIoU=0.5)} \\ \hline
Baseline & 3886.9                       & 795.9                 & 76.01 \\ 
1 & 3725.5 \textbf{{{(95.8\%)}}} & 765.4 \textbf{{{(96.2\%)}}} & 77.97 \textbf{{{($\uparrow$ 1.96)}}} \\ 
2 & 3564.1 \textbf{{{(91.7\%)}}} & 731.4 \textbf{{{(91.9\%)}}} & 78.00 \textbf{{{($\uparrow$ 1.99)}}} \\ 
3 & 3402.7 \textbf{{{(87.5\%)}}} & 704.3 \textbf{{{(88.5\%)}}} & 78.29 \textbf{{{($\uparrow$ 2.28)}}} \\ 
4 & 3241.4 \textbf{{{(83.4\%)}}} & 671.3 \textbf{{{(84.3\%)}}} & 77.86 \textbf{{{($\uparrow$ 1.85)}}} \\ 
5 & 3080.0 \textbf{{{(79.2\%)}}} & 647.1 \textbf{{{(81.3\%)}}} & 77.57 \textbf{{{($\uparrow$ 1.56)}}} \\ 
6 & 2918.6 \textbf{{{(75.1\%)}}} & 602.6 \textbf{{{(75.7\%)}}} & 77.25 \textbf{{{($\uparrow$ 1.24)}}} \\ \hline
\end{tabular}
}\label{exp:block drop large}
\vspace{-0.5cm}
\end{table}

\subsection{Implementation Details}\label{sec:implementation details}
We use the ActionFormer~\cite{zhang2022actionformer} detector for prediction, employing two Transformer-based feature extractors: VideoMAE-S with 12 blocks and VideoMAE-L~\cite{tong2022videomae} with 24 blocks, both are pre-trained on the K400 dataset~\cite{kay2017kinetics}. For the THUMOS14 dataset, the input videos are randomly truncated to a fixed length of 768 frames, while for the ActivityNet-1.3 and FineAction dataset, the input videos are resized to a fixed length of 768 frames, with the frame resolution set to $160^2$ by default for both datasets. Please refer to the supplementary for more details.

\subsection{Effectiveness of Our Progressive Block Drop}\label{exp:apply block drop method}
We compare the model trained with the progressive block drop method to the uncompressed model from three aspects. \textbf{1) Computational Complexity (MACs):} The computational complexity is calculated based on the number of multiply-accumulate operations at each layer of the model. \textbf{2) Inference Time:} The inference time measures the average duration for processing each input sample, where each sample consists of 768 frames with a resolution of $160^2$. \textbf{3) mAP:} The results, as shown in Table~\ref{exp:block drop small}, demonstrate that on both datasets, our method achieves an approximately 25\% reduction in MACs while outperforming the uncompressed model in terms of performance. This highlights the superiority of our approach, as it effectively reduces computational complexity while simultaneously enhancing the performance of the compressed model—an accomplishment that is difficult for other compression methods to achieve.

\textbf{Drop blocks on a deeper backbone.} We explore the performance of the block drop method when the depth of the feature extractor is increased from 12 to 24, i.e., when transitioning from VideoMAE-S to VideoMAE-L. The results, as shown in Table~\ref{exp:block drop large}, reveal that with a 25\% reduction in MACs, the model achieves a 1.24\% performance improvement. Notably, when the same fraction of blocks (1/4) is dropped, the performance of VideoMAE-L improves by 1.24\%, whereas VideoMAE-S only shows a modest improvement of 0.04\% (as shown in Table~\ref{exp:block drop small}). This suggests that in models with higher computational complexity, there is more redundant information, and the progressive block drop method is effective at eliminating such redundancy.

\textbf{Dropping multiple blocks at one time $vs$ Progressive dropping.}\label{sec:drop one time}
To demonstrate the necessity of adopting a multi-step progressive drop in our proposed block drop method, we conduct experiments where multiple blocks are dropped simultaneously. Specifically, we randomly select three different blocks to drop and train the model using the training method described in Section~\ref{sec:Performance Recovery}. The results, as shown in Table~\ref{exp:sim_drop}, indicate that directly dropping three blocks yields performance ranging from 68.31\% to 69.76\%, with the best result still 0.67\% lower than that of the uncompressed model. Meanwhile, using the progressive block drop method on VideoMAE-S and the THUMOS14 dataset, the model's performance after dropping the first three layers [7, 10, 6] reaches 70.47\%. Comparing this with the result of directly dropping these three layers, it is evident that compressing the model via progressive dropping achieves better performance recovery (70.47\% vs. 68.91\%), further underscoring the importance of the multi-step progressive drop. Therefore, our proposed multi-step progressive block drop is crucial for performance recovery.

\begin{table}[t]
\tabcolsep 2pt
\centering
\caption{
Comparison with randomly dropping blocks. 
Dropping three randomly selected blocks at once performs worse than progressively dropping.
}
 \vspace{-0.3cm}
\resizebox{1.\columnwidth}{!}{
\small
    \begin{tabular}{c|cccc|c}
    \hline
    Drop Block ID & [2,5,9] & [0,4,7] & [1,3,8] & [6,7,10] & [6,7,10] + ours\\ \hline
    mAP & 68.88 & 68.31 & 69.76 & 68.91 & \textbf{70.47} \\ \hline
    \end{tabular}
}
\label{exp:sim_drop}
\vspace{-0.6cm}
\end{table}

\subsection{Extended Evaluations}\label{exp:extended eval}
\vspace{-0.1cm}
\textbf{Evaluation on additional TAD architectures.} To validate the generalizability of our method, we apply it to two additional TAD architectures: the single-stage AdaTAD~\cite{liu2024end} model with a VideoMAE-S backbone and the two-stage ActionFormer~\cite{zhang2022actionformer} model with pre-extracted I3D features. Table~\ref{exp:drop_other_work} shows that in AdaTAD, pruning two blocks improves accuracy by 0.08\% and reduces MACs to 84.0\%. Similarly, the ActionFormer model gains 0.49\% in mAP while reducing MACs to 59.7\%. These results underscore our method's effectiveness in compressing diverse TAD architectures while enhancing performance.

\begin{table}[t]
\centering
\caption{Evalution of our method on AdaTAD and ActionFormer. Our method is effective across different TAD architectures.}
\vspace{-0.3cm}
\tabcolsep=1pt
\resizebox{\columnwidth}{!}{
\begin{tabular}{c|cc|cc}
\hline
\multirow{2}{*}{Drop \#Block} & \multicolumn{2}{c|}{AdaTAD~\cite{liu2024end}} & \multicolumn{2}{c}{ActionFormer~\cite{zhang2022actionformer}} \\ \cline{2-5} 
 & \multicolumn{1}{c}{mAP (tIoU=0.5)} & \multicolumn{1}{c|}{MACs (G)} & \multicolumn{1}{c}{mAP (tIoU=0.5)} & \multicolumn{1}{c}{MACs (G)} \\ \hline
Baseline & 72.43 & 324.8 & 72.65 & 45.2 \\
1 & 72.91 \textbf{($\uparrow$ 0.48)} & 298.8 \textbf{(92.0\%)} & 73.21 \textbf{($\uparrow$ 0.56)} & 37.9 \textbf{(83.8\%)} \\
2 & 72.51 \textbf{($\uparrow$ 0.08)} & 272.8 \textbf{(84.0\%)} & 73.28 \textbf{($\uparrow$ 0.63)} & 30.6  \textbf{(67.7\%)} \\
3 & 71.68 \textbf{($\downarrow$ 0.75)} & 246.9 \textbf{(76.0\%)} & 73.14 \textbf{($\uparrow$ 0.49)} & 27.0 \textbf{(59.7\%)} \\ \hline
\end{tabular}
}\label{exp:drop_other_work}
\vspace{-0.3cm}
\end{table}

\textbf{Evaluation on the fine-grained FineActions dataset.} To further corroborate the versatility of our approach, we conduct experiments on the fine-grained FineActions~\cite{liu2022fineaction} dataset using a VideoMAE-S backbone with an ActionFormer~\cite{zhang2022actionformer} head. As shown in Table~\ref{exp:fineaction}, pruning two blocks slightly improves mAP ($\uparrow$ 0.04\%), while pruning three blocks causes only a minor drop ($\downarrow$ 0.24\%). Meanwhile, MACs are reduced to 83.5\% and 75.3\%, with a notable decrease in inference time. These results demonstrate that our progressive block drop strategy effectively preserves detection accuracy while significantly reducing computational cost, even for fine-grained action detection.

\begin{table}[t]
\caption{Evaluation of our method on fine-grained FineActions~\cite{liu2022fineaction}. Our approach continues to exhibit efficacy even in scenarios involving more fine-grained actions.} 
 \vspace{-0.3cm}
\tabcolsep=3pt
\centering
\resizebox{\columnwidth}{!}{
\normalsize
\begin{tabular}{cccc}
\hline
Drop \#Block & mAP (Avg)      & MACs (G)          & \multicolumn{1}{l}{Inference Time (ms)} \\ \hline
Baseline            & 18.67         & 280.8           & 210.2                                  \\ 
1            & 18.77 \textbf{($\uparrow$ 0.10)} & 257.7 \textbf{(91.8\%)} & 197.0 \textbf{(93.7\%)}                        \\ 
2            & 18.71 \textbf{($\uparrow$ 0.04)}  & 234.6 \textbf{(83.5\%)} & 185.0 \textbf{(88.0\%)}                        \\ 
3            & 18.43 \textbf{($\downarrow$ 0.24)}  & 211.5 \textbf{(75.3\%)} & 173.0 \textbf{(82.3\%)}                               \\ \hline
\end{tabular}
}\label{exp:fineaction}
\vspace{-0.3cm}
\end{table}

\textbf{Evaluation on the natural language localization task.} Finally, to test our method beyond conventional TAD, we apply progressive block drop to the grounding head of the Mamba model~\cite{chen2024video} on Charades-STA~\cite{gao2017tall} for natural language localization. Selective pruning is performed on the grounding head. As shown in Table~\ref{exp:vg_mamba}, removing three out of six blocks reduces the parameter count to 65.1\% of the baseline while increasing performance—R1@0.3 by 2.48\% and R1@0.5 by 0.21\%. These results demonstrate our method's generalizability across tasks and models.

\begin{table}[t]
\caption{Results of pruning Mamba model~\cite{chen2024video} on Charades-STA. Our method is effective across various tasks and architectures.}
 \vspace{-0.3cm}
\tabcolsep=3pt
\centering
\resizebox{\columnwidth}{!}{
\normalsize
\begin{tabular}{cccc}
\hline
Drop \#Block & Parameters (M)  & R1@0.3         & R1@0.5 \\ \hline
Baseline & 60.5                      & 69.46          & 58.58  \\ 
1 & 53.4 \textbf{(88.4\%)}   & 72.04 \textbf{{{($\uparrow$ 2.58)}}} & 58.79 \textbf{{{($\uparrow$ 0.21)}}} \\ 
2 & 46.4 \textbf{(76.7\%)}   & 71.75 \textbf{{{($\uparrow$ 2.29)}}} & 58.92 \textbf{{{($\uparrow$ 0.34)}}} \\ 
3 & 39.3 \textbf{(65.1\%)}   & 71.94 \textbf{{{($\uparrow$ 2.48)}}} & 58.79 \textbf{{{($\uparrow$ 0.21)}}} \\ 
4 & 32.3 \textbf{(53.4\%)}   & 70.83 \textbf{{{($\uparrow$ 1.37)}}} & 57.72 \textbf{{{($\downarrow$ 0.86)}}} \\ \hline
\end{tabular}
}\label{exp:vg_mamba}
\vspace{-0.3cm}
\end{table}

\vspace{-0.1cm}
\subsection{Compatible with Pruning Techniques}\label{sec:compatibility}
\begin{figure}[t]
    \centering 
    \includegraphics[width=0.9\linewidth]{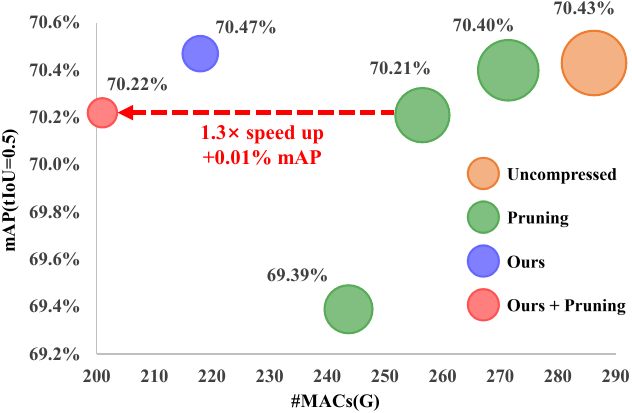}
    \vspace{-0.3cm}
    \caption{
     Comparison and compatibility with the pruning method. Bubble size represents the model's computational complexity. Combining our method with pruning enables further 1.3$\times$ acceleration, demonstrating our compatibility with pruning.
    }
    \label{fig:Pruning_+_Drop} 
    \vspace{-0.3cm}
\end{figure}

\vspace{-0.1cm}
\noindent We apply the pruning method~\cite{fang2023depgraph} to models after block drop, considering three setups: \textbf{1) pruning}, \textbf{2) our method}, and \textbf{3) our method with pruning}. As shown in Figure~\ref{fig:Pruning_+_Drop}, at nearly identical accuracy (70.22\% vs. 70.21\%), combining our method with pruning reduces MACs by 30\%, while pruning alone achieves only 10\%. This highlights our method's potential for further integration with pruning to accelerate the model. Additionally, compatibility with sparse activations is analyzed in Section~\ref{sec:sa_compatibility} of the appendix.

\begin{table}[t]
\renewcommand{\arraystretch}{1.0}
\centering
\caption{
Comparisons of inference time between pruning methods and ours under the same computational complexity on THUMOS14, VideoMAE-S backbone. Reducing depth yields better acceleration than reducing width.
}
 \vspace{-0.3cm}
\resizebox{1.0\columnwidth}{!}{
\normalsize
\begin{tabular}{c|ccc}
\hline
Model        & MACs (G) & Inference Time (ms) & Speedup \\ \hline
Uncompressed & 286.3    & 104.9               & 1.00$\times$   \\ 
Pruning      & 218.2    & 96.6                & 1.09$\times$   \\
Ours         & 218.0    & 81.2                & 1.29$\times$   \\ \hline
\end{tabular}
}
\label{exp:Pruning and ours}
 \vspace{-0.5cm}
\end{table}

\subsection{Comparison of Inference Time with Pruning}\label{sec:Inference Time}
 We test the inference times of models compressed via pruning and our method when both achieved the same reduction in MACs. The inference time is measured as the average duration per input sample, each consisting of 768 frames with a resolution of $160^2$. As shown in Table~\ref{exp:Pruning and ours}, under similar MACs (218.2G $vs$ 218.0G), pruning achieves a $1.09\times$ speedup, while the block drop method achieves a $1.29\times$ speedup. This indicates that compressing the model by reducing network depth is more effective for acceleration than reducing network width because shallow and wide models are more hardware-friendly architectures~\cite{liu2020autocompress}.

\vspace{-0.2cm}
\section{Conclusion}
\vspace{-0.1cm}
In this work, we have proposed a progressive block drop method to compress TAD models from the depth perspective. This method consists of two main steps: 1) The block selection evaluator automatically selects blocks to be dropped based on their importance; 2) Parameter-efficient learning combined with three alignment techniques is used to recover performance. On the THUMOS14 and ActivityNet-1.3 datasets, we have achieved a reduction of about 25\% in computational cost while improving the pruned model's performance above that of the uncompressed model. We have further validated its generalizability across TAD architectures, fine-grained datasets, and other tasks, demonstrating compatibility with existing compression techniques. Additionally, we have demonstrated the compatibility of our method with other compression techniques through experiments. We hope that our work paves the way for further advancements in the compression and acceleration of TAD models.

{\flushleft \bf Acknowledgements}. This work was partially supported by National Natural Science Foundation of China (NSFC) under Grants 62202311, the Guangdong Basic and Applied Basic Research Foundation under Grants 2023A1515011512, Excellent Science and Technology Creative Talent Training Program of Shenzhen Municipality under Grant RCBS20221008093224017,
the Shenzhen Natural Science Foundation (the Stable Support Plan Program) under Grant 20220809180405001, Key Scientific Research Project of the Department of Education of Guangdong
Province 2024ZDZX3012.

{
    \small
    
    \bibliographystyle{ieeenat_fullname}

    \bibliography{main}
    
}

\appendix 
\setcounter{figure}{0}
\renewcommand\thefigure{\Alph{figure}}
\setcounter{table}{0}
\renewcommand\thetable{\Alph{table}}
\renewcommand\thesubfigure{(\alph{subfigure})} 
\maketitlesupplementary
\captionsetup{font={normalsize}}

In the supplementary material, we provide more details and more experimental results of our work. We organize the supplementary into the following sections.
\begin{itemize}
    \item In Section~\ref{Sec:Imp_detail}, we provide a detailed description of the hyperparameter settings used in our experiments.

    \item In Section~\ref{Sec:more_result_setting}, we present detailed results on which layers are dropped at each iteration using the block drop method.

    \item In Section~\ref{Sec:detad}, we employ the DETAD tool to conduct a comprehensive analysis of model performance across a broader range of metrics.  
    
    \item In Section~\ref{sec:standard deviation}, we investigate the data stability of our method through repeated experiments.

    \item In Section~\ref{sec:sa_compatibility}, we explore the compatibility of the block drop method with sparse activations.

    \item In Section~\ref{sec:abliation study}, we further validate the effectiveness of our method through five ablation experiments.
\end{itemize}

\section{Implementation Details}\label{Sec:Imp_detail}
We use PyTorch 2.3.1 and 4 A800 GPUs for our experiments. The learning rate of LoRA is grid-searched within the range of 1e-5 to 1e-3, while the parameters of the other backbone components remain frozen. The hidden layer dimension of LoRA is fixed at one-quarter of the input dimension of the attention block. Other experimental details follow those described in~\cite{liu2024end}.

\section{Detailed Results of the Dropped Blocks}\label{Sec:more_result_setting}
As discussed in Section 5.3, we applied our method to the THUMOS14 and ActivityNet-1.3 datasets. Here, we provide more detailed experimental results and configurations, including the specific blocks removed from each layer and the learning rate settings. As shown in Table~\ref{exp:detal_setting}, the dropped blocks vary across models, demonstrating that the block drop selection evaluator can flexibly adapt to the characteristics of each model when selecting blocks to remove.

\begin{table}[t]
\caption{
The detailed dropped blocks for the experiments in Section 5.3. The dropped blocks vary across different models, indicating that our method is capable of adapting to the specific characteristics of each model.
}
\vspace{-0.2cm}
\centering
\normalsize
\begin{tabular}{c|c|c}
\hline
Dataset         & Backbone     & Drop Blocks  \\ \hline
\multirow{10}{*}{THUMOS14}       & \multirow{4}{*}{VideoMAE-S}   & {[}7{]}              \\ 
                                 &                               & {[}7,10{]}         \\
                                 &                               & {[}7,10,6{]}       \\
                                 &                               & {[}7,10,6,3{]}     \\ \cline{2-3} 
                                 & \multirow{6}{*}{VideoMAE-L}   & {[}20{]}               \\
                                 &                               & {[}20,1{]}         \\
                                 &                               & {[}20,1,4{]}       \\
                                 &                               & {[}20,1,4,7{]}     \\
                                 &                               & {[}20,1,4,7,6{]}   \\
                                 &                               & {[}20,1,4,7,6,2{]} \\ \hline
\multirow{4}{*}{ActivityNet-1.3} & \multirow{4}{*}{VideoMAE-S}   & {[}6{]}               \\
                                 &                               & {[}6,3{]}          \\
                                 &                               & {[}6,3,8{]}        \\ 
                                 &                               & {[}6,3,8,11{]}     \\ \hline
\end{tabular}
\label{exp:detal_setting}
\end{table}

\section{Quantitative Analysis with DETAD~\cite{alwassel2018diagnosing}}\label{Sec:detad}

\begin{figure}[t]
    \centering 
    \includegraphics[width=1.\linewidth]{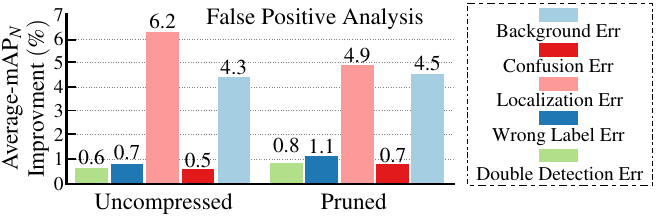}
    \vspace{-0.6cm}
    \caption{Quantitative analysis between uncompressed and pruned model. The pruned model enhances localization performance while diminishing classification performance.}
    \label{Fig.DETAD_FP} 
    \vspace{-0.4cm}
\end{figure}

In order to assess the impact of our progressive block drop on detection performance, we conducted a detailed quantitative analysis of the pruned model using the DETAD~\cite{alwassel2018diagnosing}. Our analysis focuses on the false positive predictions before and after compression. As illustrated in Figure~\ref{Fig.DETAD_FP}, pruning notably improves localization, reducing the combined background and localization errors from 6.2+4.3 to 4.9+4.5. However, this improvement comes at the expense of classification accuracy, with the classification error increasing from 0.7 to 1.1. 

\begin{figure}[t]
    \centering 
    \includegraphics[width=\linewidth]{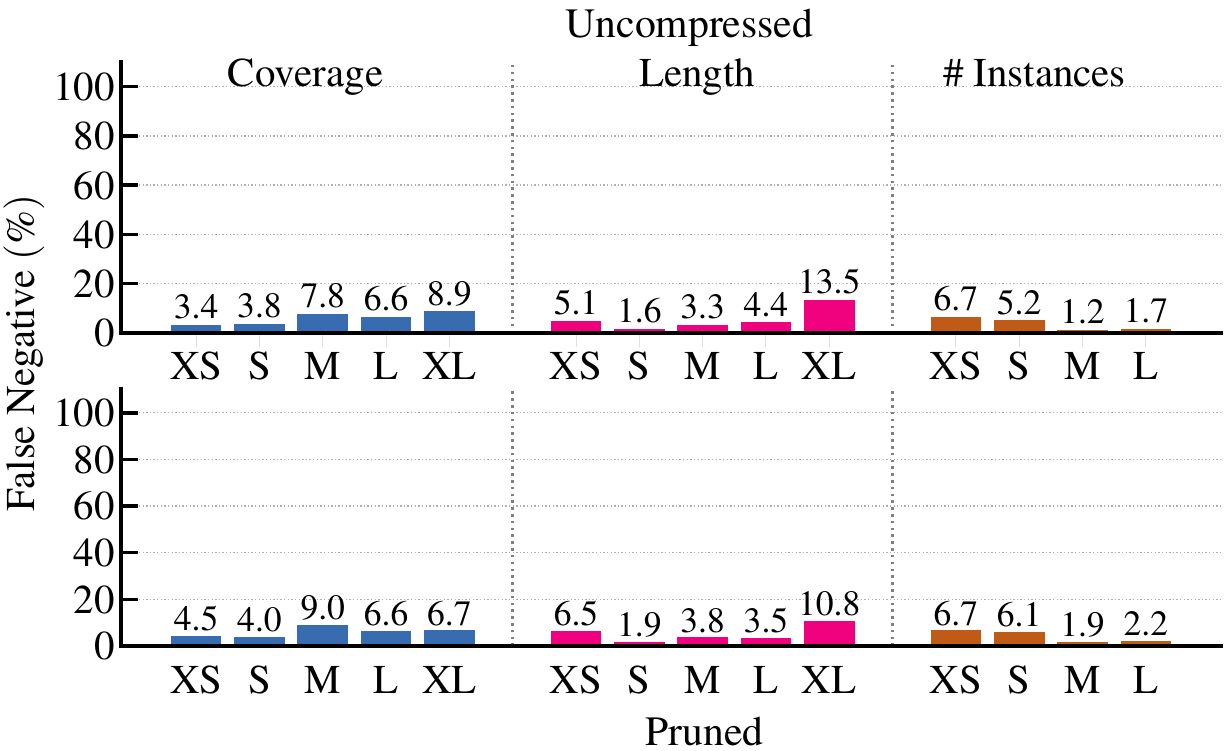}
    \vspace{-0.7cm}
    \caption{
    Comparison of false negative analysis between uncompressed and pruned models.
    }
    \label{Fig.thumos_fn} 
\end{figure}

Furthermore, we used the DETAD tool to examine false negative predictions based on three metrics: \textbf{1) Coverage:} The proportion of an instance's duration relative to the video's total duration, divided into five categories. \textbf{2) Length:} The duration of an instance in seconds, categorized into five groups from short to extra-long. \textbf{3) Number of Instances:} The total count of same-class instances within a video, grouped into four categories. As shown in Figure~\ref{Fig.thumos_fn}, the model with progressive block drop improved the detection of long-duration actions, reducing omission rates for actions with XL coverage from 8.9\% to 6.7\% and for XL length from 13.5\% to 10.8\%. This demonstrates the potential of our compression method in analyzing long-duration actions.

\section{Standard deviation of Results.}\label{sec:standard deviation}
To compare the performance stability of our progressive block drop method with the random block drop approach. We conduct the experiment using three different random seeds to evaluate the performance of both methods: (1) the baseline approach where blocks are randomly dropped all at once, and (2) our progressive block drop method. The evaluation is performed on two datasets: THUMOS14 and ActivityNet-1.3. The experimental results are summarized in Table~\ref{exp:random_seed}. On THUMOS14, our method achieves a mean accuracy of 70.57\% (±0.18), while randomly dropping blocks results in lower accuracy (69.06\% and 68.41\%). On ActivityNet-1.3, our method shows a slight improvement, with an accuracy of 37.74\% (±0.03), compared to 37.57\% for the random drop approach. This confirms the statistical significance and robustness of our method.

\begin{table}[h]
\caption{Comparison between ours and randomly dropping blocks. The results show that our method consistently achieves higher accuracy and more stable performance.}
\vspace{-0.2cm}
\tabcolsep=3pt
\centering{
\resizebox{\columnwidth}{!}{
\begin{tabular}{c|cc|c}
\hline
Drop Blocks ID  & {[}2,5,9{]}       & {[}0,4,7{]}       & Ours              \\ \hline
THUMOS14        & 68.41 (±0.06) & 69.06 (±0.01) & 70.57 (±0.18) \\ 
ActivityNet-1.3 & 37.57 (±0.08) & 37.57 (±0.01) & 37.74 (±0.03) \\ \hline
\end{tabular}
}}
\label{exp:random_seed} 
\vspace{-0.4cm}
\end{table}

\section{Combined with Sparse Activations.}\label{sec:sa_compatibility}
\noindent Activation sparsity is another promising approach for achieving model acceleration~\cite{kurtz2020inducing}. In theory, it leverages the high density of zero values to accelerate subsequent computations by using sparse matrices. Given a model compressed using our method, we replace the GeLU activation function in each block with the ReLU function, following~\cite{mirzadehrelu}. We then conduct experiments by fine-tuning this model using the alignment training method described in Section 4.2. As shown in Table~\ref{exp:Our + Sparsity}, after replacing the activation functions in all blocks with ReLU, the performance changed from 70.47\% to 70.38\%, which is essentially unchanged, while the activation sparsity increased from 0.18\% to 83.42\%. Therefore, it is entirely feasible to further accelerate the model's inference speed by combining specialized hardware and sparse matrix computation algorithms. This also demonstrates the strong scalability of our proposed progressive block drop method.

\begin{table}[t]
\centering
\caption{
Results of our compatibility with sparse activation-based acceleration methods. Our pruned model can be further accelerated by sparse activation.
}
 \vspace{-0.2cm}
\tabcolsep 13pt
\small
\begin{tabular}{c|cc}
\hline
Model & Ours & Ours + Sparse Activation  \\ \hline
mAP & 70.47 & 70.38 \\ 
Sparsity & 0.18\% & \textbf{83.42\%} \\ \hline
\end{tabular}
\label{exp:Our + Sparsity}
\vspace{-0.2cm}
\end{table}

\section{Ablation Studies}\label{sec:abliation study}
\subsection{Different Choices of Block Drop Criteria}\label{sec:different metrics}
We compare different importance metrics to select which blocks to drop. We employed three metrics: \textbf{1) Train Loss}: the average loss over all data when training the subnet after dropping blocks. \textbf{2) MSE}: perform inference on the training set and compute the feature differences before and after each block. The average MSE over all input data is used as the evaluation metric. \textbf{3) mAP}: performance of the subnet evaluated on the training set. From Table~\ref{exp:different evaluation metrics}, when using mAP to select blocks to drop, the performance on both datasets surpasses that of the uncompressed model (at least $\uparrow$ 0.02\%). While the train loss metric does not exceed the uncompressed model on the ActivityNet-1.3 dataset, it is relatively close ($\downarrow$ 0.04\%) and can outperform the uncompressed model on the THUMOS14 dataset. However, MSE performs worse on both datasets (at least $\downarrow$ 0.50\%). Therefore, we choose mAP as the evaluation metric due to its more stable performance across different datasets.

\begin{table}[t]
\renewcommand{\arraystretch}{1.0}
\centering
\caption{
Comparison of three metrics in the block drop selection evaluator. Training loss and mAP benefits more on block selection. 
}
 \vspace{-0.2cm}
\resizebox{1.0\columnwidth}{!}{
\normalsize
\begin{tabular}{cc|ccc}
\hline
 \multirow{2}{*}{Dataset} & \multirow{2}{*}{Uncompressed} & \multicolumn{3}{c}{Drop Metric} \\ \cline{3-5} 
                 &       & Train Loss     & mAP            & MSE   \\ \hline
 THUMOS14        & 70.43 & \textbf{70.88} & 70.47          & 69.37 \\ 
 ActivityNet-1.3 & 37.75 & 37.71          & \textbf{37.77} & 37.25 \\ \hline
\end{tabular}
}\label{exp:different evaluation metrics}
\vspace{-0.2cm}
\end{table}

\subsection{Ablation on the Effect of Fine-Tuning After Block Dropping}\label{sec:effect of FT}
Our experiments underscore the critical importance of fine-tuning following block dropping. In our ablation studies, as illustrated in Table~\ref{exp:skip_ft}, pruning three blocks without any fine-tuning led to a marked decline in mAP ($\downarrow$ 5.14\%). In contrast, when fine-tuning was applied, the mAP was effectively restored to 70.47\%. This recovery highlights that fine-tuning is essential to bridge the domain gap between the pre-trained model and the target TAD dataset, thereby mitigating the adverse effects of structural modifications.

\begin{table}[t]
\centering
\caption{Effect of fine-tuning (FT) after block dropping. FT after block pruning is essential for restoring performance.}
\vspace{-0.2cm}
\tabcolsep=1pt
\setlength{\tabcolsep}{3pt}
\resizebox{\columnwidth}{!}{
\small
\begin{tabular}{c|cccc}
\hline
Drop \#Block & Baseline & 1 & 2 & 3 \\ \hline
without FT & 70.43 &
{69.21 \textbf{($\downarrow$ 1.22)}} &
{67.13 \textbf{($\downarrow$ 3.30)}} &
{65.29 \textbf{($\downarrow$ 5.14)}} \\ 
with FT & 70.43  &
  71.06 \textbf{($\uparrow$ 0.63)} &
  71.37 \textbf{($\uparrow$ 0.94)}&
  70.47 \textbf{($\uparrow$ 0.04)}\\ \hline
\end{tabular}
}\label{exp:skip_ft}
\end{table}

\subsection{Ablation on the Impact of LoRA}\label{sec:impact of LoRA}
To further enhance the adaptation process while reducing computational overhead, we integrate Low-Rank Adaptation (LoRA) into our fine-tuning strategy. As presented in Table~\ref{exp:lora}, the incorporation of LoRA achieves comparable—and in some cases superior—accuracy relative to full fine-tuning, yet requires significantly fewer trainable parameters. For example, when pruning two blocks, the LoRA-augmented approach attained an mAP improvement of 0.94\% with only 2.68M tunable parameters, in contrast to 17.48M when employing full fine-tuning. These results validate that LoRA not only minimizes the computational cost of fine-tuning but also contributes positively to performance, confirming its practical utility in our framework.

\begin{table}[t]
\caption{Comparison between LoRA and full fine-tuning. Param refers to the trainable parameters within the backbone. The use of LoRA technology can effectively reduce computational costs while benefiting model performance.}
\vspace{-0.2cm}
\tabcolsep=2pt
\resizebox{\columnwidth}{!}{
{
\small
\begin{tabular}{c|cc|cc}
\hline
\multirow{2}{*}{Drop \#Block} & \multicolumn{2}{c|}{Full Fine-tuning} & \multicolumn{2}{c}{LoRA} \\  
                              & mAP             & Param (M)         & mAP     & Param (M)  \\ \hline
Baseline                             & 70.43          &        20.86            & 70.43   &   3.11           \\
1                             & 70.67 \textbf{($\uparrow$ 0.24)}          & 19.17              & 71.06 \textbf{($\uparrow$ 0.63)}   & 2.90        \\
2                             & 70.75 \textbf{($\uparrow$ 0.32)}          & 17.48              & 71.37 \textbf{($\uparrow$ 0.94)}   & 2.68        \\
3                             & 70.42 \textbf{($\downarrow$ 0.01)}          & 15.79              & 70.47 \textbf{($\uparrow$ 0.04)}   & 2.47       \\ \hline
\end{tabular}
}\label{exp:lora}}
\end{table}

\subsection{Ablation on Selective Fine-Tuning}
We further explored the efficiency of our adaptation strategy by freezing the detection head and all network components except the LoRA modules within the backbone. This approach resulted in a modest mAP decrease of 1.57\% (from 70.47\% to 68.90\%) when three blocks were pruned. Although selective fine-tuning of only the LoRA blocks considerably reduces the number of trainable parameters, the performance drop suggests that jointly fine-tuning both the backbone and the detection head is preferable for optimal performance recovery. These findings indicate that while freezing the detection head can offer additional computational savings, a comprehensive fine-tuning strategy remains essential to fully recover the performance.

\begin{table}[t]
   \tabcolsep 4pt
    \renewcommand{\arraystretch}{1.0}
    \centering
    \caption{
    Ablation on different alignment losses on THUMOS14.
    The combination of both alignment losses yields the best result. 
    }\vspace{-0.2cm}
{
\small
\begin{tabular}{cc|cccc}
\hline
\multicolumn{1}{c}{\multirow{2}{*}{Alignment Level}} & Feature    &       &    \checkmark   &       &    \checkmark   \\ \cline{2-6}
\multicolumn{1}{c}{}                                 & Prediction &       &       &    \checkmark   &   \checkmark    \\ \hline
\multicolumn{2}{c|}{mAP (tIoU=0.5)}                                & 69.25 & 70.23 & 70.39 & \textbf{70.47} \\ \hline
\end{tabular}
}\label{Tab.align_loss}
\end{table}

\subsection{Ablation on Loss Functions for Alignment}\label{sec:different loss}
To investigate the effect of different alignment losses on the performance recovery, we design several ablation experiments: \textbf{1) no alignment loss}; \textbf{2) action recognition and localization alignment}; \textbf{3) cross-depth feature alignment}. The experimental results in Table~\ref{Tab.align_loss}, demonstrate that the best performance is achieved when both alignment techniques are used simultaneously. This indicates that combining these two losses more effectively guides the model to learn in a way that is closer to the uncompressed model.

\end{document}